\documentclass[letterpaper]{article} %
\usepackage{aaai25}  %
\usepackage{times}  %
\usepackage{helvet}  %
\usepackage{courier}  %
\usepackage[hyphens]{url}  %
\usepackage{graphicx} %
\usepackage{natbib}  %
\usepackage{caption} %
\usepackage{algorithm}
\usepackage{algorithmic}

\usepackage{graphicx}
\usepackage{booktabs}
\usepackage{graphicx, amsmath, amssymb, caption, subcaption, multirow, overpic, textpos, pifont, adjustbox}

\RequirePackage{xspace}
\makeatletter
\DeclareRobustCommand\onedot{\futurelet\@let@token\@onedot}
\def\@onedot{\ifx\@let@token.\else.\null\fi\xspace}

\def\ie{\emph{i.e}\onedot} 
 
\def\etc{\emph{etc}\onedot}

\def\etal{\emph{et al}\onedot}
\makeatother
\usepackage{newfloat}
\usepackage{listings}
\DeclareCaptionStyle{ruled}{labelfont=normalfont,labelsep=colon,strut=off} %
\floatstyle{ruled}
\newfloat{listing}{tb}{lst}{}
\floatname{listing}{Listing}
\title{TinySAM: Pushing the Envelope for Efficient Segment Anything Model}
\author{
	Han Shu\textsuperscript{\rm 1,2},
	Wenshuo Li\textsuperscript{\rm 2}, 
	Yehui Tang\textsuperscript{\rm 2}, 
	Yiman Zhang\textsuperscript{\rm 2}, \\
	Yihao Chen\textsuperscript{\rm 2}, 
	Houqiang Li\textsuperscript{\rm 1}, 
	Yunhe Wang\textsuperscript{\rm 2$*$}, 
	Xinghao Chen\textsuperscript{\rm 2}\thanks{Corresponding authors.}
}
\affiliations{
    \textsuperscript{\rm 1}University of Science and Technology of China\\
    \textsuperscript{\rm 2}Huawei Noah’s Ark Lab\\
    \{han.shu, xinghao.chen, yunhe.wang\}@huawei.com

}

\usepackage{bibentry}

\begin{document}

\maketitle

\begin{abstract}
Recently segment anything model (SAM) has shown powerful segmentation capability and has drawn great attention in computer vision fields. Massive following works have developed various applications based on the pre-trained SAM and achieved impressive performance on downstream vision tasks. 
However, SAM consists of heavy architectures and requires massive computational capacity, which hinders the further application of SAM on computation constrained edge devices. To this end, in this paper we propose a framework to obtain a tiny segment anything model (TinySAM) while maintaining the strong zero-shot performance. We first propose a full-stage knowledge distillation method with hard prompt sampling and hard mask weighting strategy to distill a lightweight student model. We also adapt the post-training quantization to the prompt-based segmentation task and further reduce the computational cost. Moreover, a hierarchical segmenting everything strategy is proposed to accelerate the everything inference by $2\times$ with almost no performance degradation. With all these proposed methods, our TinySAM leads to orders of magnitude computational reduction and pushes the envelope for efficient segment anything task. Extensive experiments on various zero-shot transfer tasks demonstrate the significantly advantageous performance of our TinySAM against counterpart methods. Codes are
available at \url{https://github.com/xinghaochen/TinySAM} and \url{https://gitee.com/mindspore/models/tree/master/research/cv/TinySAM}.
\end{abstract}

\section{Introduction}
Object segmentation is an important and foundational task in computer vision fields. Extensive visual applications such as object localization and verification rely on accurate and fast object segmentation. Tremendous prior works have focused on segmentation tasks which include semantic segmentation~\cite{fcnseg,segmenter}, instance segmentation~\cite{bolya2019yolact,liu2018path} and panoptic segmentation~\cite{maskformer,kirillov2019panoptic}. Recently, Kirillov~\etal~\cite{SAM} introduce a powerful segment anything model (SAM), together with a massive segmentation dataset SA-1B that contains over 1 billion masks on 11 million images. With the strong capability to segment objects with arbitrary shapes and categories, SAM has become a foundation framework for many downstream tasks such as object tracking~\cite{cheng2023segment}, image inpainting~\cite{yu2023inpaint} and 3D vision~\cite{cen2023segment3D} \etc. Moreover, the powerful zero-shot segmentation ability of SAM has benefited research area with less data like medical imaging~\cite{ma2023segment}.   

Although SAM has achieved impressive performance on downstream vision tasks, complicated architecture and huge computational cost make SAM difficult to be deployed on resource constrained devices. The inference time of SAM model for a 1024$\times$1024 image could take up to $2$ seconds on a modern  GPU~\cite{fastsam}. Some recent attempts have tried to obtain a more computation efficient segment anything model. For example, MobileSAM~\cite{mobilesam} tries to replace the heavy component of image encoder with a lightweight architecture of TinyViT~\cite{tinyvit}. However, it only accesses the image encoder network with a decoupled knowledge distillation strategy by training the compact image encoder network with the supervision of image embeddings from the teacher network. This partial training strategy inevitably causes  performance decay without the supervision of final mask prediction. FastSAM~\cite{fastsam} transfers the segment anything task to an instance segmentation task with only one foreground category with YOLOv8~\cite{yolov8_ultralytics1}. To fulfill the function of prompt-based segmentation, FastSAM applies a post-process strategy together with the instance segmentation network. However, this reformulated framework could not achieve comparable performance as SAM on downstream zero-shot tasks.

\begin{figure*}[tb]
	\centering
	\includegraphics[width=0.99\textwidth]{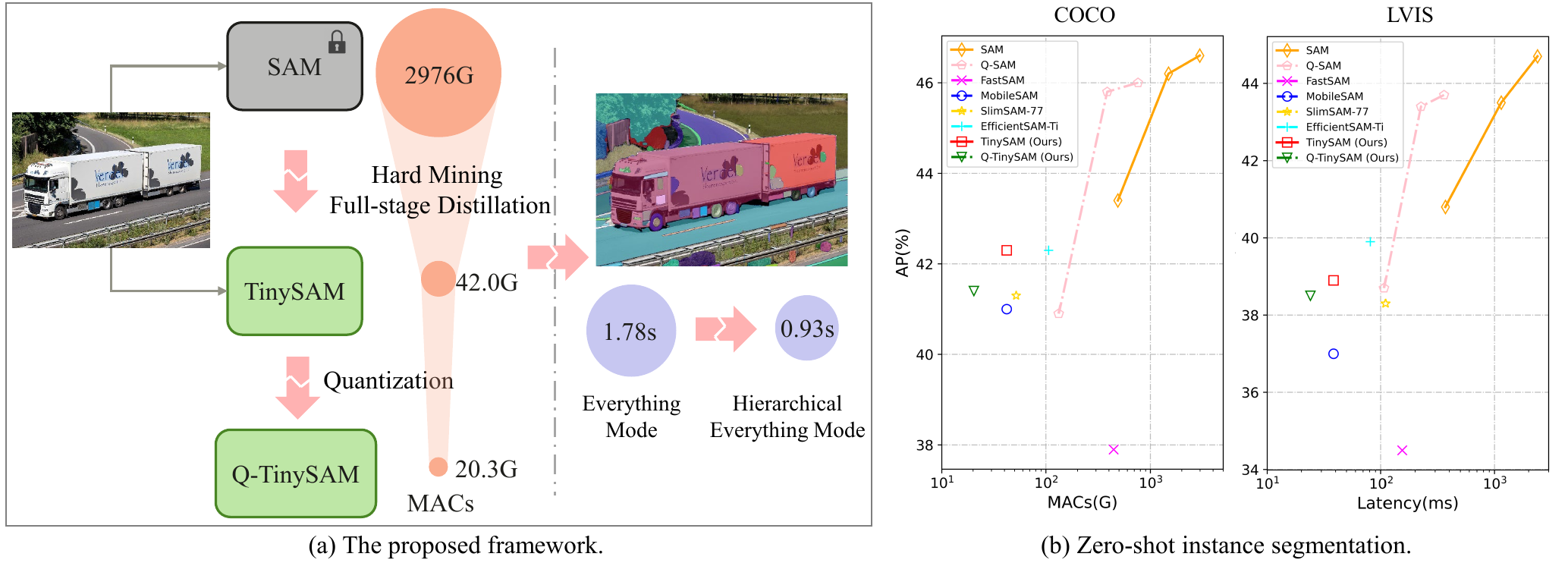}
	\vspace{-2mm}
	\caption{(a)~The overall framework of our proposed method. Consisting the modules of the hard mining full-stage knowledge distillation, the post training quantization and the hierarchical everything inference,  the computation cost is down-scaled by magnitudes. (b) The proposed TinySAM can save considerable computation cost while maintaining the performance. The latency is tested with TensorRT on NVIDIA T4 GPU.}
	\label{framework}
	\vspace{-6mm}
\end{figure*}

To further push the envelope for efficient segment anything model, in this paper we propose a full framework to obtain TinySAM that greatly reduces the computational cost while maintaining the zero-shot segmentation ability to maximum extent.  Specifically, we propose a hard mining full-stage knowledge distillation method to improve the capability of the compact student network. The student network is distilled in an end-to-end manner with the supervision of teacher network from different network stages. A mask-weighted distillation loss is proposed to efficiently transfer the information from teacher to student through massive various SA-1B masks. Besides, an online hard prompt sampling strategy is proposed to make the distillation process attend more to hard examples and thus improves the final performance.  
We also adapt the post-training quantization to the prompt-based segmentation task and further reduce the computational cost.
Moreover, we find that it takes tremendous computational cost for segmenting everything in an image since massive masks have to be generated from grid prompt points.
To this end, a hierarchical segmenting everything strategy is proposed to accelerate the everything inference by $2\times$ with almost no performance degradation. 
With all these proposed methods, our TinySAM leads to orders of magnitude computational reduction and pushes the envelope for efficient segment anything task. For example, TinySAM can achieve 100$\times$ acceleration for segment anything task compared with the original SAM. Extensive experiments on various zero-shot transfer tasks demonstrate the significantly advantageous performance of our TinySAM against counterparts.

\section{Related Work}
\label{sec:2_relatedwork}

\subsection{Segment Anything Model}
Recently proposed segment anything model~(SAM)~\cite{SAM} proves its generalization on object segmentation and downstream vision tasks. SAM consists of three subnetworks, \emph{i.e.}, image encoder, prompt encoder and mask decoder. The image encoder is a heavy vision transformer-based network~\cite{vit}, which extracts the input image into image embedding. The prompt encoder is designed to encode input points, boxes, arbitrary-shaped masks and free-form text with positional information. The geometric prompt and text prompt are processed with different networks. The mask decoder, which contains a two-way transformer, takes the output of image encoder and prompt encoder to generate the final mask prediction. Together with the proposed SA-1B dataset, which contains 11 million high-resolution images and more than 1 billion high-quality segmentation masks, SAM shows impressive high quality segmentation ability for objects of any category and shape. Moreover, SAM demonstrates powerful generalization on zero-shot downstream vision tasks including edge detection, object proposal, instance segmentation and text-to-mask prediction. Due to the flexible prompt mode and high quality segmentation capability, SAM has been regarded as a foundation model for vision applications. However, SAM, especially the image encoder network, consists of large parameters and requires high computation capacity for deployment. Therefore, it is not easy to apply SAM on edge devices with constrained resources. The compression and acceleration of SAM becomes an important research topic~\cite{fastsam,mobilesam,slimsam}.

\subsection{Knowledge Distillation}
Hinton~\emph{et al.}~\cite{hinton2015distilling} propose the knowledge distillation method to supervise the training of lightweight student network via the output of teacher network. Since then knowledge distillation has been an important approach to improve the performance of compact networks during training process. Knowledge distillation methods can be roughly divided into two categories,~\emph{i.e.} distillation for network outputs~\cite{hinton2015distilling} and for intermediate features~\cite{romero2014fitnets}. Majority of research of knowledge distillation methods have focused on image classification task ~\cite{Park_2019_CVPR,peng2019correlation,dong2023improving,li2022spatial}. Subsequent works~\cite{chen2017kd,liu2019structured,guo2021distilling,chen2020optical,Deng_2019_ICCV} propose knowledge distillation methods for high-level computer vision tasks such as object detection and semantic segmentation. 
Zhang~\etal\cite{mobilesam} propose to use the distillation method to obtain an efficient segment anything model (MobileSAM). However, MobileSAM only accesses the image encoder network with the supervision of corresponding image embeddings from original SAM. This partial distillation strategy could cause considerable performance decay since there is no guidance of mask-level information for lightweight student network from either teacher network or labeled data. 

\subsection{Quantization}

Model quantization is also one of the commonly used model compression methods, which quantizes weights or activations from higher bit-width to lower bit-width to reduce both storage requirements and computational complexity with limited accuracy degradation. There are two types of model quantization methods, quantization-aware training (QAT)~\cite{choi2018pact,esser2019learned} and post-training quantization (PTQ) ~\cite{choukroun2019low}. QAT methods require a labeled training dataset and extensive training cost, while PTQ methods only need a small unlabeled calibration dataset and thus are more efficient. 
Many prior PTQ methods~\cite{liu2023pd,nagel2020up} have proposed to search for appropriate quantization parameters for convolutional neural networks. As vision transformers~\cite{vit,liu2021swin} achieve remarkable performance on various visual tasks, recent works~\cite{liu2021post,yuan2022ptq4vit,tai2023tsptq,li2022q} investigate how to apply post-training quantization for vision transformers and have achieved good performance with 8-bit quantization configuration. However, there is rare exploration for quantization of prompt-based segmentation task, especially for segment anything models.

\section{Methodology}

\subsection{Overview of TinySAM}
This paper proposes a framework to get a highly efficient SAM, as described in Figure~\ref{framework}. Firstly, we introduce a hard mining full-stage knowledge distillation specifically designed for SAM. To further activate the distillation process, the proposed hard mask weighting and hard prompt sampling strategy are utilized to mine the essential knowledge from the teacher network to the student network. Secondly, a post-training quantization method is adapted to prompt-based segmentation task and applied to the lightweight student network. Thirdly, a hierarchical everything inference mode is designed for segmenting everything task, which can avoid massive redundant computation only with negligible accuracy loss and speedup the inference time by $2\times$.  

\begin{figure}
	\centering
	\includegraphics[width=0.99\linewidth]{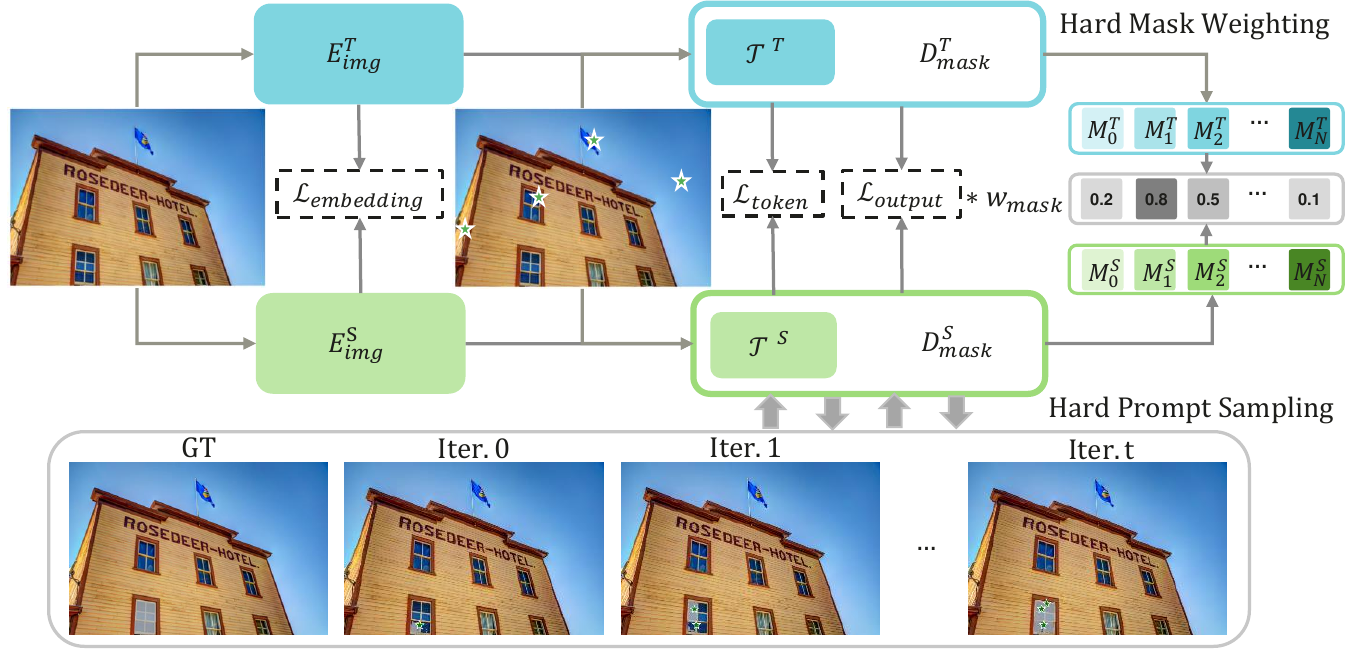}
	\vspace{0mm}
	\caption{The framework of the hard mining full-stage knowledge distillation. For the massive masks of SA-1B dataset, we design the hard prompt sampling for prompts and hard mask weighting for distillation loss. For sampling process, the stars represent sampling point with different iterations. With the increase of iterations, the sampling region is more closed to the edge of the target mask, which makes the prompt relatively harder for student network to learn. Moreover, according to the gap between student and teacher network, different weight is assigned to each mask when calculating the distillation loss.}
	\label{distillation}
	\vspace{-6mm}
\end{figure} 

\subsection{Hard Mining Full-Stage Knowledge Distillation}\label{sec:sec_distillation}
SAM consists of three subnetworks, \emph{i.e.} image encoder, prompt encoder and mask decoder. The image encoder network is based on vision transformer~\cite{vit} and consumes great computation cost. Inspired by MobileSAM~\cite{mobilesam}, we use the lightweight TinyViT~\cite{tinyvit} to replace the original heavy image encoder network. Considerable performance decay exists for this simple substitution. Therefore, we propose a hard mining full-stage knowledge distillation strategy to guide the lightweight image encoder during learning procedure from multiple knowledge levels. 

Besides the conventional loss between the predicted results and ground-truth labels, we introduce multiple distillation losses on different stages as described in Figure~\ref{distillation}. Specifically, we select several nodes of teacher network to guide the learning of student network from multiple level of knowledge. Firstly, we choose the output feature of image encoder, \emph{i.e.} image embedding, as a distillation information. Image embedding concentrates the information from input image, which is the fundamental knowledge during the prediction. For an input image of $\mathit{I}$, the distillation loss function for image embedding can be expressed as,
\begin{equation}\small
	\mathcal{L}_{embedding} = \mathcal{L} (\mathit{E}_{img}^{T} (\mathit{I}), \mathit{E}_{img}^{S} (\mathit{I})),  
	\label{eq:embloss}
\end{equation} 
where $\mathit{E}_{img}^{S}$ and $\mathit{E}_{img}^{T}$  denote the image encoder for student and teacher network, respectively.
Since image level information does not directly relate to the mask prediction, features more close to the final output are essential for this segmentation task. Naturally, the final output of the teacher network is chosen to be a distillation point.  The output distillation loss $\mathcal{L}_{output}$ can be described as,
\begin{equation}\small
	\small
	\mathcal{L}_{output} = \mathcal{L} (\mathit{D}_{mask}^{T}(\mathit{E}_{img}^{T} (\mathit{I}),\textit{q}), \mathit{D}_{mask}^{S}(\mathit{E}_{img}^{S} (\mathit{I}),\textit{q})),  
	\label{eq:output} 
\end{equation}
where  $\mathit{D}_{mask}^{S}$ and $\mathit{D}_{mask}^{T}$ are  mask decoders for student and teacher, respectively. \textit{q} denotes the query of the mask decoder, which is the concatenation of prompt embedding and output tokens.
Since the structure of SAM is rather complicated, the previously mentioned two distillation losses could be inconsistent and thus hard for lightweight student to learn. We further propose to distill the output tokens from the two-way transformer of the mask decoder, which interacts information from prompt embedding and image embedding. It captures the target mask information in a more abstract way. The corresponding distillation losses $\mathcal{L}_{token}$ can be described as,    
\begin{equation}\small
	\mathcal{L}_{token} = \mathcal{L} (\mathcal{T}^{T}(\mathit{E}_{img}^{T} (\mathit{I}),\textit{q}), \mathcal{T}^{S}(\mathit{E}_{img}^{S} (\mathit{I}),\textit{q})),  
	\label{eq:tokenloss} 
\end{equation} 
where $\mathcal{T}^{S}$ and  $\mathcal{T}^{T}$  are the two-way transformer module of mask decoder and $\mathcal{L}$ denotes the loss function. We empirically find that the numerical values of feature difference could make the conventionally used MSE loss ($\ell_2$ distance) too small to be well optimized. Thus we use $\ell_1$ distance function instead. The overall distillation loss function $\mathcal{L}_{distill}$ can be expressed as,
\begin{equation}\small
	\mathcal{L}_{distill} = \alpha* \mathcal{L}_{embedding} +\beta * \mathcal{L}_{token} + \gamma * \mathcal{L}_{output},    
\end{equation} 
where $\alpha$, $\beta$, $\gamma$ represent the hyper-parameters for each distillation loss. The total training loss is a linear combination of distillation loss, ground truth loss for mask prediction $\mathcal{L}_{mask}$ and IoU prediction $\mathcal{L}_{ious}$, where $\mathcal{L}_{mask}$ is a combination of focal loss~\cite{focal} and dice loss~\cite{dice}, $\mathcal{L}_{ious}$ is $\ell_1$ loss function between predicted IoUs and calculated IoUs.
\begin{equation}\small
	\mathcal{L}_{total} = \mathcal{L}_{distill} +  \mathcal{L}_{mask} +  \mathcal{L}_{ious}.
\end{equation} 
{\noindent\bf Hard Mask Weighting.} To make the knowledge distillation more effective, we design a hard mask weighting strategy when calculating the losses. There is an observation that masks could be extremely various in a single image of SA-1B dataset since the fine-grained granularity and no semantic constraints. As shown in Figure~\ref{distillation}, segmenting the flag with complex boundary could be difficult while segmenting the rectangular window with high contrast color could be easy.  The hard mask should reasonably be assigned with larger weight for student to learn. Specifically, we calculate the gap of student and teacher network output to indicate the mask hardness $\mathcal{H}_{i}$.
\begin{equation}\small
	\mathcal{H}_{i} = \mathrm{sigmoid}(\frac{\mathrm{IoU}({M}_{i}^{T},{M}_{i}^{GT})}{\mathrm{IoU}({M}_{i}^{S},{M}_{i}^{GT})+\epsilon}-1),
\end{equation}
where ${M}_{i}^{T}$, ${M}_{i}^{S}$, ${M}_{i}^{GT}$ represent the mask prediction of student network, the mask prediction of teacher network and the ground truth for $i$th mask, respectively. Thus the distillation loss could be updated with 
\begin{equation}\small
	\mathcal{L}_{distill}^{*} = \alpha* \mathcal{L}_{embedding} +\beta * \mathcal{L}_{token} + \gamma * \sum_{i=1}^{N}\mathcal{H}_{i}* \mathcal{L}_{output}^{i}.   
\end{equation} 

{\noindent\bf Hard Prompt Sampling.} Generally, random sampling from labeled training data could be adopted to generate the prompts to drive the end-to-end training of prompt-based mask prediction network as SAM. To further ease the learning process of the distillation between teacher and lightweight student network, we propose a hard prompt sampling strategy, which makes the training samples concentrate in the difficult area for prediction. Taking points prompt as an example, points $P_0$ are initially sampled inside the labeled mask region $M_{gt}$. These initial points are fed into the network with input image to get the predicted mask region $M_0$. Then we sample the prompt points from the difference set of $M_{gt}$ and $M_0$, and we conduct the procedure iteratively. The $(\mathit{i}+1)$-th round sampling points $P_i$ are sampled from the difference set of  $M_{gt}$ and $M_{i}$, \ie
\begin{equation}\small
	P_{i+1} \in M_{gt} -M_{i}, i=0,1,2,...
\end{equation}
where
\begin{equation}\small
	M_{i}= \mathit{D}_{mask} (\mathit{E}_{prompt}(P_i),\mathit{E}_{img}(\mathit{I}) ). 
\end{equation}  
When applied on the training process, the $i$-th iteration is random sampled from $0$ to $9$, which makes the difficulty of sampled prompts in a constrained range. The bottom of Figure~\ref{distillation} shows the location change of the sampling prompts with iterations, the green stars denote the sampled point prompts with online hard prompt sampling strategy. With more iterations, the sampling points are more close to the edge region of the ground truth mask.

\subsection{Quantization}{\label{method:quant}}
Quantization aims to project floating point tensor $x$ to $b$-bit integer tensor $x_q$ with a scaling factor $s$. The uniform symmetric quantization could be formulated as follows,
\begin{equation}\small
	x_q=Q(b, s)=\textrm{clip}(\textrm{round}(\frac{x}{s}), -2^{b-1}, 2^{b-1}-1). 
\end{equation}
For a matrix multiplication $O=AB$, it can be quantized with two scaling factors $s_A$ and $s_B$, and the quantized matrix is denoted as $\hat{O}=\hat{A}\hat{B}$. The metric for measuring the distance between $\hat{O}$ and $O$ is vitally important for optimizing $\hat{A}$ and $\hat{B}$. Following the successful practice of quantization methods in image classification models~\cite{tai2023tsptq,yuan2022ptq4vit,frantar2022gptq,wu2020easyquant}, we perform hessian guided metric as the distance to solve the scaling factors, which is more consistent with task loss. Different from classification tasks, the prompt-based segmentation task of SAM outputs segmentation predictions which contains fine-grained masks. Thus we use the Kullback-Leible (KL) divergence of masks and IoUs as the task loss and use some calibration data to calculate the hessian matrix, the task loss is formulated as,
\begin{equation}\small %
	L=\textrm{KL}(\hat{y}_{pred}, y_{pred}) + \textrm{KL}(\hat{y}_{iou}, y_{iou}),
\end{equation}
where $y_{pred}$ and $y_{iou}$ are the outputs of the floating point model, $\hat{y}_{pred}$ and $\hat{y}_{iou}$ are the outputs after quantization.

After specifying the distance metric, we could solve $s_A$ and $s_B$ as an alternate iterative grid search problem. With calibration data we get the maximum value of $A$ and $B$, which is $A_{max}$ and $B_{max}$ respectively, and use two parameters $\theta_l$ and $\theta_u$ to specify the search range for $s_A$ and $s_B$, $[\theta_l \frac{A_{max}}{2^{b-1}}, \theta_u \frac{A_{max}}{2^{b-1}}]$ and $[\theta_l \frac{B_{max}}{2^{b-1}}, \theta_u \frac{B_{max}}{2^{b-1}}]$. These two search ranges are linearly divided into $n$ candidate options separately.  $\hat{A}$ and $\hat{B}$ are optimized in an alternate manner. %

The input of matrix multiplication after softmax is unevenly distributed at both ends of the interval [0,1], while the feature after GELU varies greatly between the positive and negative ranges. These two circumstances go far from the assumption of uniform quantization, \ie, the activation in neural networks obeys Gaussian distribution. The violation will result in high quantization error. Thus we split feature into two groups and use two scaling factors to reduce the quantization error. %

\begin{figure}[t]
	\centering
	\includegraphics[width=0.95\linewidth]{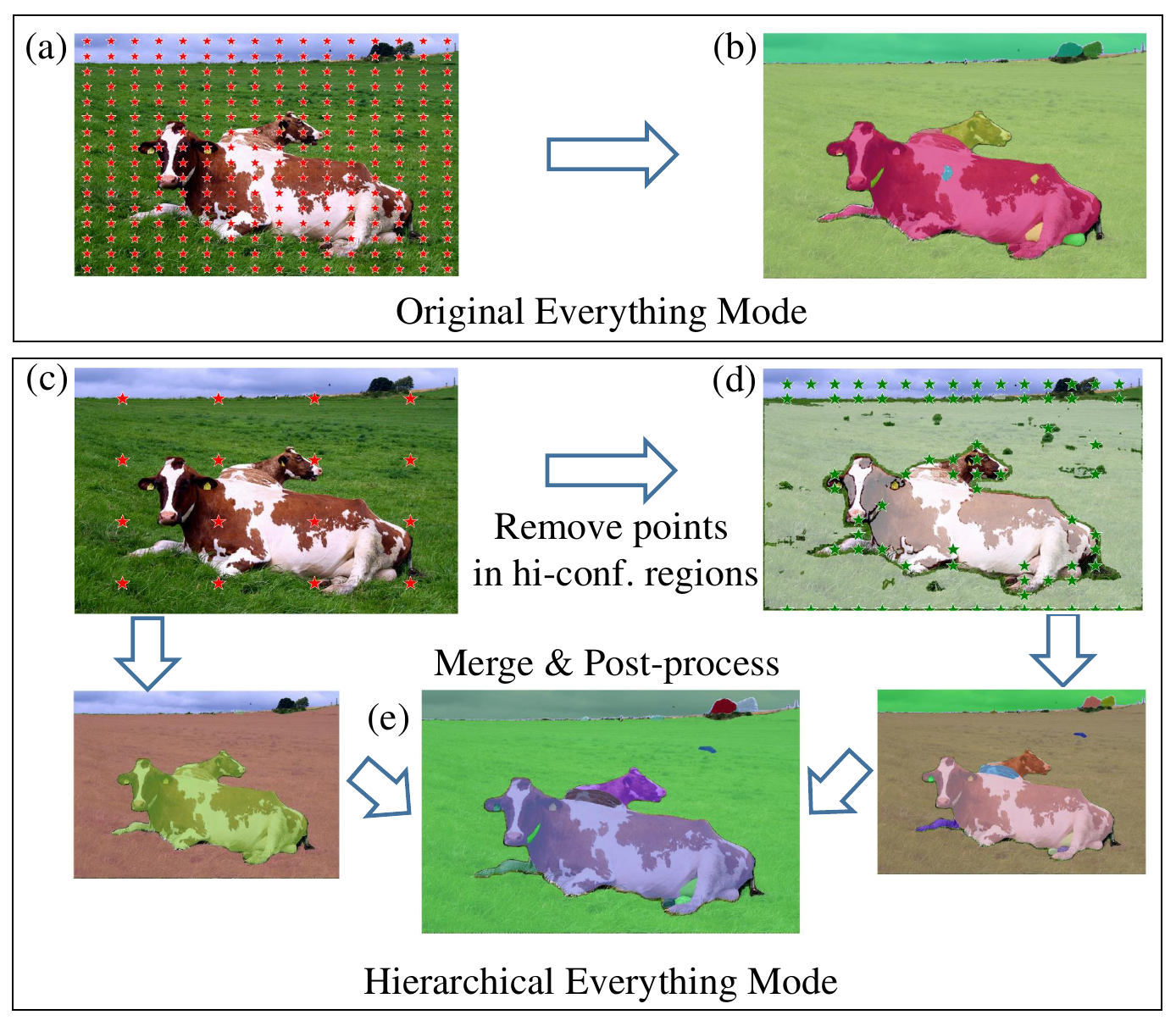}
	\vspace{-2mm}
	\caption{Comparison between our hierarchical strategy and the original strategy. (a) Points sampling (take \textit{points\_per\_side=16} as an example) of original everything mode.  (b) Segmentation results of original strategy. (c) First step of our hierarchical strategy, only $1/16$ points are sampled. (d) Get high confidence area from (c) and ignore points in this area. The high confidence area is shown as white mask. (e) Segmentation results of our hierarchical strategy.}
	\label{fig:everything-mode}
	\vspace{-6mm}
\end{figure}

\subsection{Hierarchical Segmenting Everything}\label{sec:meth:ethinfer}

SAM proposes an automatic mask generator which samples points as a grid to segment everything. However, we find that dense point grid leads to over fine-grained segmentation results and also occupies massive computing resources. On the one hand, for a complete object, too many sampling points may cause slightly different parts of the object to be incorrectly segmented as separate masks. On the other hand, since the image encoder has been largely shrunk by the proposed method, the time cost of everything mode inference is mainly in the mask decoder part. For the default setting of SAM automatic mask generator, it samples $32\times 32=1024$ points as the prompts, which means the mask decoder is inferred by $1024$ times. It costs $16$ms for image encoder and $894$ms for mask decoder on a single V100 GPU.

To reduce the time cost of everything mode, we propose a hierarchical mask generating method. The comparison between our hierarchical strategy and the original one is shown in Figure~\ref{fig:everything-mode}. Different from original everything mode, in the first step we only use $1/4$ points in each side so the total points is $1/16$ of the original settings, as shown in Figure~\ref{fig:everything-mode}(c).  Then we infer the prompt encoder and mask decoder with these prompts and get the results. 

Then we filter out some masks with confidence exceeding a threshold $\tau$, and mark the corresponding regions as areas that could be considered as final predictions. For these areas, since they are considered as the segmentation results of instances with high confidences, there is no need to re-generate point prompts. Thus we sample points as the same density with original setting but ignore points in the above area. As shown in Figure~\ref{fig:everything-mode}(d), most points on the grass and body of the front cow are ignored. Meanwhile, the points on the back cow and the sky are kept to be further segmented. Specifically, the back cow is incorrectly segmented as the same object with the front cow in the initial round. This strategy can avoid redundant cost of inference time and over fine-grained segmentation of the object. Then we utilize the point prompts sampled in the second round to get the mask predictions. Finally, the results of these two round are merged and post-processed to get the final masks. More than $50\%$ points are ignored by our method thus brings in significant latency reduction.

\section{Experiments}
\subsection{Implementation Details}
We utilize the TinyViT-5M~\cite{tinyvit} as the lightweight student image encoder and SAM-H as the teacher model, following prior work~\cite{mobilesam}.  1\% of SA-1B dataset is used as the training data for full-stage distillation. We adopt Adam optimizer and train the student network for 8 epochs. For each iteration, we sample 64 prompts according to hard prompt sampling strategy. To accelerate the distillation process, the image embeddings from the teacher network have been computed and stored in advance. Therefore, the heavy image encoder of teacher network is not necessary to compute repeatedly during training time. 
For post training quantization, we set $\theta_l=0.01, \theta_u=1.2, n=100, rounds=3$ for iterative search. We calibrate quantized model on SA-1B dataset using 8 images. We conduct zero-shot evaluation on downstream tasks like instance segmentation and point prompt segmentation. Following the suggestions by SAM~\cite{SAM}, the multi-output mode is adopted and the final mask prediction is the one with highest IoU prediction. %

\subsection{Zero-Shot Instance Segmentation}\label{sec:eval:instseg}
For zero-shot instance segmentation task, we strictly follow the experimental settings of SAM and use the object detection results of ViTDet-H~\cite{li2022exploring} as the box prompt for instance segmentation. We evaluate the zero-shot instance segmentation task for models on the benchmark of COCO~\cite{lin2014microsoft} dataset and LVIS v1~\cite{gupta2019lvis}. We compare our TinySAM with different variants of SAM~\cite{SAM}, and also with prior efficient models like FastSAM~\cite{fastsam}, MobileSAM~\cite{mobilesam}, EfficientSAM~\cite{xiong2023efficientsam} and SlimSAM~\cite{slimsam}. As shown in Table~\ref{tab:instance_segmentation}, the proposed TinySAM obtained superior performance when compared with prior methods. Specifically, our TinySAM outperforms FastSAM~\cite{fastsam} in terms of MACs and instance segmentation accuracy, \ie, about $4\%$ AP improvement with only $9.5\%$ MACs and $25\%$ latency.
With the same computational cost, our TinySAM also achieves $1.3\%+$ AP on COCO dataset than MobileSAM~\cite{mobilesam} and $1.9\%+$ AP on LVIS v1 dataset, respectively. With similar performance on COCO dataset, TinySAM is $2\times$ faster than EfficientSAM~\cite{xiong2023efficientsam}.
Our W8A8 quantized variant of TinySAM (Q-TinySAM) also obtains competitive performance across different methods. Specifically, Q-TinySAM achieves $0.1\%+$ AP on COCO and $0.2\%+$ on LVIS v1 dataset than SlimSAM~\cite{slimsam}, with only $39\%$ MACs and $21.8\%$ latency. Visual results on COCO validation set and LVIS dataset are shown in the appendix. Our proposed TinySAM captures more clear and smooth boundaries compared with other efficient variants of SAM.
\begin{table*}[t]
	\centering
	\small
	\setlength{\tabcolsep}{9.pt}
	\renewcommand{\arraystretch}{1.0}
	\begin{tabular}{l|c|c|cccc|cccc}%
		\toprule\hline
		\multirow{2}{*}{} & \multirow{2}{*}{} &\multirow{2}{*}{} & \multicolumn{4}{c}{COCO} & \multicolumn{4}{c}{LVIS v1}\\
		Method & MACs & Lat.(ms) & AP & AP\textsuperscript{S} & AP\textsuperscript{M} & AP\textsuperscript{L} & AP & AP\textsuperscript{S} & AP\textsuperscript{M} & AP\textsuperscript{L} \\
		\midrule
		ViTDet-H~\cite{li2022exploring} & -& - & 51.0 & 32.0 & 54.3 & 68.9 & 46.6 & 35.0 & 58.0 & 66.3\\
		\multicolumn{9}{l}{\emph{zero-shot transfer methods (segmentation module only):}} \\
		SAM-H~\cite{SAM} %
		& 2976G & 2392 & 46.6 & 30.8 & 51.0 & 61.7 & 44.7 & 32.5 & 57.6 & 65.5\\
		SAM-L~\cite{SAM} 
		& 1491G &1146 & 46.2 & 30.2 & 50.1 & 60.5 & 43.5 & 31.1 & 56.3 & 65.1\\
		SAM-B~\cite{SAM}
		& 487G  &368.8 & 43.4 & 28.5 & 45.5 & 53.4 & 40.8 & 29.1 & 52.8 & 60.7 \\
		FastSAM~\cite{fastsam} %
		& 443G  &153.6 & 37.9 & 23.9 & 43.4 & 50.0 & 34.5 & 24.6 & 46.2 & 50.8 \\
		EfficientSAM-Ti~\cite{xiong2023efficientsam}
		& 106G  &81.0 & 42.3 & 26.7 & 46.2 & 57.4 & 39.9 & 28.9 & 51.0 & 59.9 \\
		SlimSAM-77~\cite{slimsam}
		& 51.7G  & 110 & 41.3 & 25.7  &  44.9 & 57.4 & 38.3 & 26.7 & 49.7  & 59.0  \\
		MobileSAM~\cite{mobilesam}
		& 42.0G &38.4 & 41.0 & 24.4 & 44.5 & 58.6 & 37.0 & 24.7 & 47.8 & 59.1 \\
		\hline
		\bf TinySAM (Ours)
		& \bf 42.0G &\bf 38.4 & \bf 42.3 & 26.3 & 45.8 & 58.8 & \bf 38.9 & 27.0 & 50.3 & 60.2 \\
		\bf Q-TinySAM (Ours)
		&\bf 20.3G &\bf 24.0 & \bf 41.4 & 25.6 & 45.1 & 57.9 & \bf 38.5 & 26.6  & 49.8  & 59.8 \\
		\hline\bottomrule
	\end{tabular}
	\vspace{-2mm}
	\caption{Zero-shot instance segmentation results on COCO and LVIS v1 dataset. Zero-shot transfer methods are prompted with the detection boxes from fully-supervised ViTDet model. TinySAM and quantized Q-TinySAM demonstrate advantageous performance on average precision. The latency is tested on NVIDIA T4 GPU.}
	\label{tab:instance_segmentation}
	\vspace{-3mm}
\end{table*}

\vspace{-2mm}
\begin{figure*}[t]
	\centering
	\includegraphics[width=0.99\linewidth]{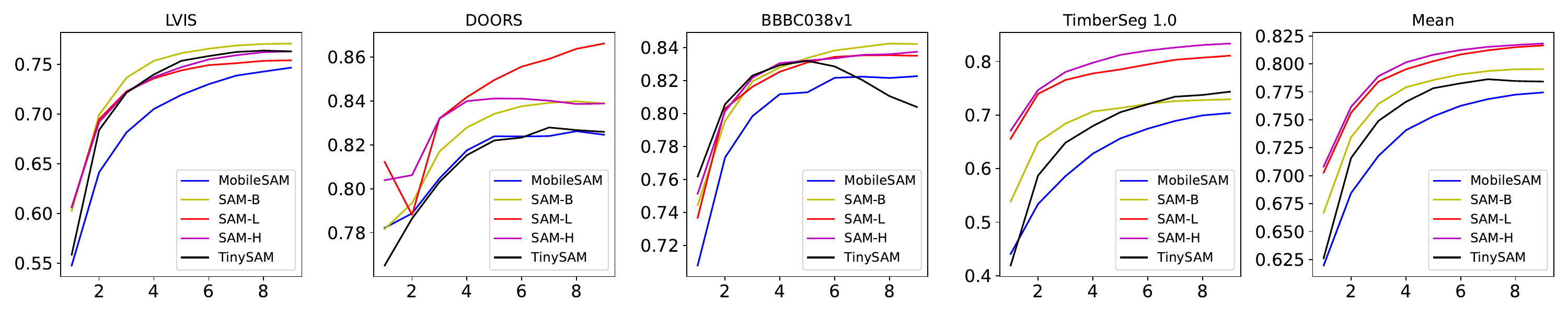}
	\vspace{-6pt}
	\caption{Results of zero-shot points valid mask evaluation. X-axis represents the number of prompts points and Y-axis represents the mIoU across all masks. The proposed TinySAM outperforms MobileSAM and achieves results close to SAM-B.}
	\vspace{-5mm}
	\label{fig:points-prompt-results}
\end{figure*}

\subsection{Zero-shot Points Valid Mask Evaluation}\label{sec:eval:points}
In this section, we evaluate the performance of our TinySAM for segmenting an object from several points as the prompts. We use the same points selection metric as previous work~\cite{SAM, gupta2019lvis}, which calculates the distance transform of false positive and false negative masks, and then sample points at a maximal value. We calculate the mIoU of each dataset to evaluate the performance of different models.

\begin{table}[t]
	\centering
	\setlength{\tabcolsep}{9.pt}
	\renewcommand{\arraystretch}{0.9}
	\small
	\begin{tabular}{l|c|c|c}
		\toprule
		Strategy    & Model & mIoU & Time (s) \\ 
		\midrule
		Original & MobileSAM & 0.5963     & 1.6719               \\
		\bf Hierarchical (Ours) & MobileSAM & 0.5958     & 0.8462               \\
		\hline
		Original & SAM-H & 0.7047     & 2.4549               \\
		\bf Hierarchical (Ours) & SAM-H & 0.7055     & 1.3537               \\  
		\hline
		Original & TinySAM & 0.6137     & 1.7790               \\
		\bf Hierarchical (Ours) & TinySAM & 0.6061     & 0.9303               \\ 
		\bottomrule
	\end{tabular}
	\vspace{-2mm}
	\caption{Comparison of original point grid strategy and our hierarchical strategy. Evaluation on the first 100 images of COCO val2017 set.}
	\label{table:everything-mode-coco}
	\vspace{-6mm}
\end{table}

We choose a subset of total 23 datasets used in~\cite{SAM} for efficient evaluation, which contains BBBC038v1~\cite{bbbc038v1}, DOORS~\cite{doors}, TimberSeg~\cite{timberseg} and LVIS~\cite{gupta2019lvis}. To make fair comparison, we follow the settings of SAM~\cite{SAM} to sample the images and masks, and the first $N$ masks in the corresponding split are used in the evaluation. 

The evaluation results are shown in Figure~\ref{fig:points-prompt-results}. Our TinySAM outperforms MobileSAM~\cite{mobilesam} significantly on LVIS and TimberSeg dataset and obtains similar performance on DOORS dataset. Moreover, TinySAM achieves better results on BBBC038v1 when fewer points are utilized as prompts. We also report the mean IoU of all four datasets, as shown in the right of Figure~\ref{fig:points-prompt-results}. The proposed TinySAM achieves higher mIoU than MobileSAM and obtains close performance to that of SAM-B.

\begin{figure}[t]
	\centering
	\includegraphics[width=0.99\linewidth]{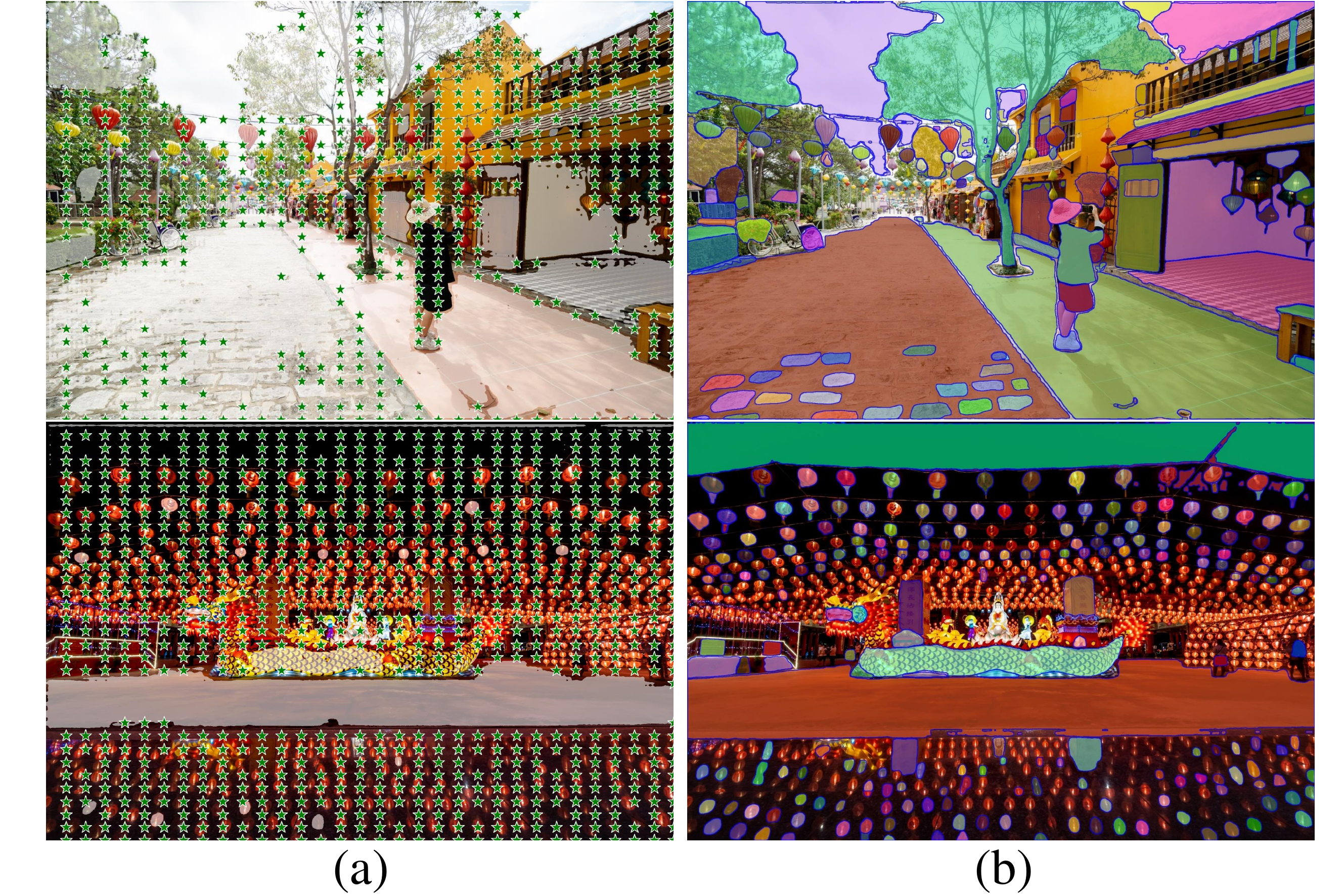}
	\vspace{-2mm}
	\caption{Visualization for the process hierarchical everything strategy. (a) shows the intermediate result of high-confidence regions after 1st sparse prompt points with white mask and remained 2nd dense prompt points with green stars. (b) shows the final segmentation result and the small objects can be accurately segmented.}
	\vspace{-6mm}
	\label{figs:hie-mid-res}
\end{figure}

\subsection{Everything Mode Acceleration}
We evaluate our proposed hierarchical everything inference strategy on COCO validation set. Latency benchmarks are conducted on a single NVIDIA V100 GPU for everything mode. We sample 100 images with the least \emph{img\_id} from val2017 and conduct everything mode inference on these images. The threshold values used in the everything mode are all kept the same as default. The results are shown in Table~\ref{table:everything-mode-coco}. We apply the same threshold and stability score on the same model evaluated with different strategies to make a fair comparison, but they can be different between these models. Our hierarchical strategy achieves comparable results compared with original $32\times 32$ points grid strategy while the cost of inference time is reduced by about $50\%$. Figure~\ref{figs:hie-mid-res} shows the intermediate visual results of the hierarchical strategy. We can see that the 1st round of sparse inference has segmented and removed the large objects, the remained points focus more on the small objects. This self-adaptive hierarchical strategy efficiently reduces the computation redundancy and maintains the high accuracy. More
visual results are shown in the appendix.

\subsection{Ablation Studies}
In this section, we conduct ablation studies of the proposed method on zero-shot instance segmentation task on COCO validation dataset. The experimental setting is the same as described in zero-shot instance segmentation. 

\noindent\textbf{Impacts of different modules.} We first evaluate the effects of different modules, \ie, full-stage knowledge distillation loss, hard prompt sampling, hard mask weighting and post quantization, respectively. As shown in Table~\ref{tab:ablation_total},  utilizing our proposed full-stage distillation strategy improve the performance from $40.7\%$ to $41.4\%$. Incorporated with the online hard prompt sampling strategy, our method could obtain $0.5\%$ AP gain. With the hard mask weighting loss, the performance can further increase to $42.3\%$. Using post-training quantization results in $0.9\%$ AP degradation but greatly reduces the computational cost.  

\begin{table}[t]
	\centering
	\setlength{\tabcolsep}{10.pt}
	\renewcommand{\arraystretch}{1.0}
	\small
	\begin{tabular}{cl|c}
		\toprule
		Ind. & Settings & AP (\%)\\
		\midrule
		0 &Baseline & 40.7 \\
		1& + Knowledge Distillation Loss  & 41.4 \\
		2& + Hard Prompt Sampling &  41.9 \\ 
		3& + Hard Mask Weighting & \bf 42.3 \\ 
		\midrule
		4&  + Quantization & 41.4 \\ 
		\bottomrule		
	\end{tabular}\vspace{-2mm}
	\caption{Effect of distillation loss, online hard prompt sampling and quantization respectively, evaluated on zero-shot instance segmentation on COCO validation dataset.}
	\label{tab:ablation_total}
	\vspace{-1mm}
\end{table}

\begin{table}[t]
	\vspace{0mm}
	\centering
	\setlength{\tabcolsep}{9.pt}
	\renewcommand{\arraystretch}{0.9}
	\small
	\begin{tabular}{cccc}
		\toprule
		Embedding Loss & Token Loss & Output Loss & AP (\%) \\
		\midrule
		-         &     -       &     \checkmark        &   41.6      \\
		\checkmark         &     -       &     \checkmark        &   41.7       \\
		\checkmark         &     \checkmark       &     \checkmark        &    41.9       \\
		\checkmark         &     \checkmark       &     \checkmark(HMW)        &    \bf 42.3       \\
		\bottomrule
	\end{tabular}
	\vspace{-2mm}
	\caption{Ablation study on combinations of knowledge distillation losses for zero-shot instance segmentation on COCO val set.}
	\label{tab:ablation_kd}
	\vspace{-1mm}
\end{table}
\begin{table}[t]
	\centering
	\setlength{\tabcolsep}{9.pt}
	\renewcommand{\arraystretch}{0.9}
	\small
	\begin{tabular}{l|c|c|c}
		\toprule
		Points per side 1st/2nd  & Thresh. $\tau$  & mIoU & Time (s) \\ 
		\midrule
		4/16 & 8.5 &   0.5521  &     0.3571          \\
		\bf{8/32} & \bf{8.5} & \bf{0.6061}     & \bf{0.9303}               \\ 
		10/32 & 8.5 &  0.6078   &   1.2774           \\
		8/32  & 7.0 &   0.6018   &      0.8154          \\  
		8/32 & 10.0 &  0.6067 &  1.1819 \\
		32/- &  -- & 0.6137     & 1.7790               \\
		\bottomrule
	\end{tabular}
	\vspace{-2mm}
	\caption{Ablation on point density and threshold for hierarchical strategy.}
	\label{table:hi_strategy}
	\vspace{-2.5mm}
\end{table}
\begin{table}[t]
	\centering
	\setlength{\tabcolsep}{9.pt}
	\renewcommand{\arraystretch}{0.9}
	\small
	\begin{tabular}{l|cc}%
		\toprule
		Model & AP (\%) & MACs (G)  \\
		\midrule
		MobileSAM &  41.0 &  42.0 \\
		+ W8A8 & 39.8 &  20.28 \\
		+ W6A6 & 36.3 &  18.45 \\
		\bf TinySAM (Ours) &  42.3 & 42.0 \\
		\bf +  W8A8 &  41.4 &  20.28 \\  %
		\bf +  W6A6 &  39.0 &  18.45 \\
		\bottomrule
	\end{tabular}\vspace{-2mm}
	\caption{Ablation study for different bit width of quantization for zero-shot instance segmentation on COCO val set.}
	\label{tab:quantization_table}
	\vspace{-2.5mm}
\end{table}

\noindent\textbf{Impacts of different distillation losses.} For detailed full-stage knowledge distillation process, we investigate the necessity of the proposed three-level distillation from the teacher network.
Table~\ref{tab:ablation_kd} shows the ablation results with different combinations of distillation losses. The output distillation loss takes important part since it is close to the supervision information and the similarity with teacher network directly reflects in the evaluation metric. Token loss and embedding loss both prove to be beneficial since they are related to key nodes of teacher network, which reflects the image-level information and the interaction of prompts with the image, respectively. Hard mask weighting for output loss can further boost the performance.

\noindent\textbf{Point density and threshold for hierarchical strategy.} In Table~\ref{table:hi_strategy}, we conduct ablation study with different settings of point density and high-confidence mask threshold $\tau$. More points and higher threshold $\tau$ lead to more precise results but longer inference time. The point density of 2nd round is more sensitive compared to the 1st one. Considering both accuracy and efficiency, the setting in bold is a good balance and used for other experiments of everything inference.

\noindent\textbf{Different bits for quantization.} 
We here explore the influence of different bit width. Table~\ref{tab:quantization_table} reports the average precision on COCO dataset. From the results, we can conclude that quantization to $8$-bit results in only slight performance drop. We also demonstrate the performance by further reducing the quantization bit width to $6$.

\section{Conclusion}
In this paper, we propose a framework to push the envelope for segment anything task and obtain a highly efficient model named TinySAM. We first propose a full-stage knowledge distillation method with hard mask weighting and hard prompt sampling strategy to distill a lightweight student model. We also adapt the post-training quantization to the prompt-based segmentation task and further reducing the computational cost. Moreover, a hierarchical segmenting everything strategy is proposed to accelerate the everything inference by $2\times$ with almost no performance degradation. With all these proposed methods, our TinySAM leads to orders of magnitude computational reduction and push the envelope for efficient segment anything task. Extensive experiments on various zero-shot transfer tasks demonstrate the significantly advantageous performance of our TinySAM against counterpart methods. We hope the proposed TinySAM brings beneficial perspective for designing a highly efficient segment anything model.

\twocolumn[{%
	\renewcommand\twocolumn[1][]{#1}%
	\vspace{2em}
	\maketitle
	\vspace{0em}
	
	\begin{center}
		\centering
		\includegraphics[width=1.0\textwidth]{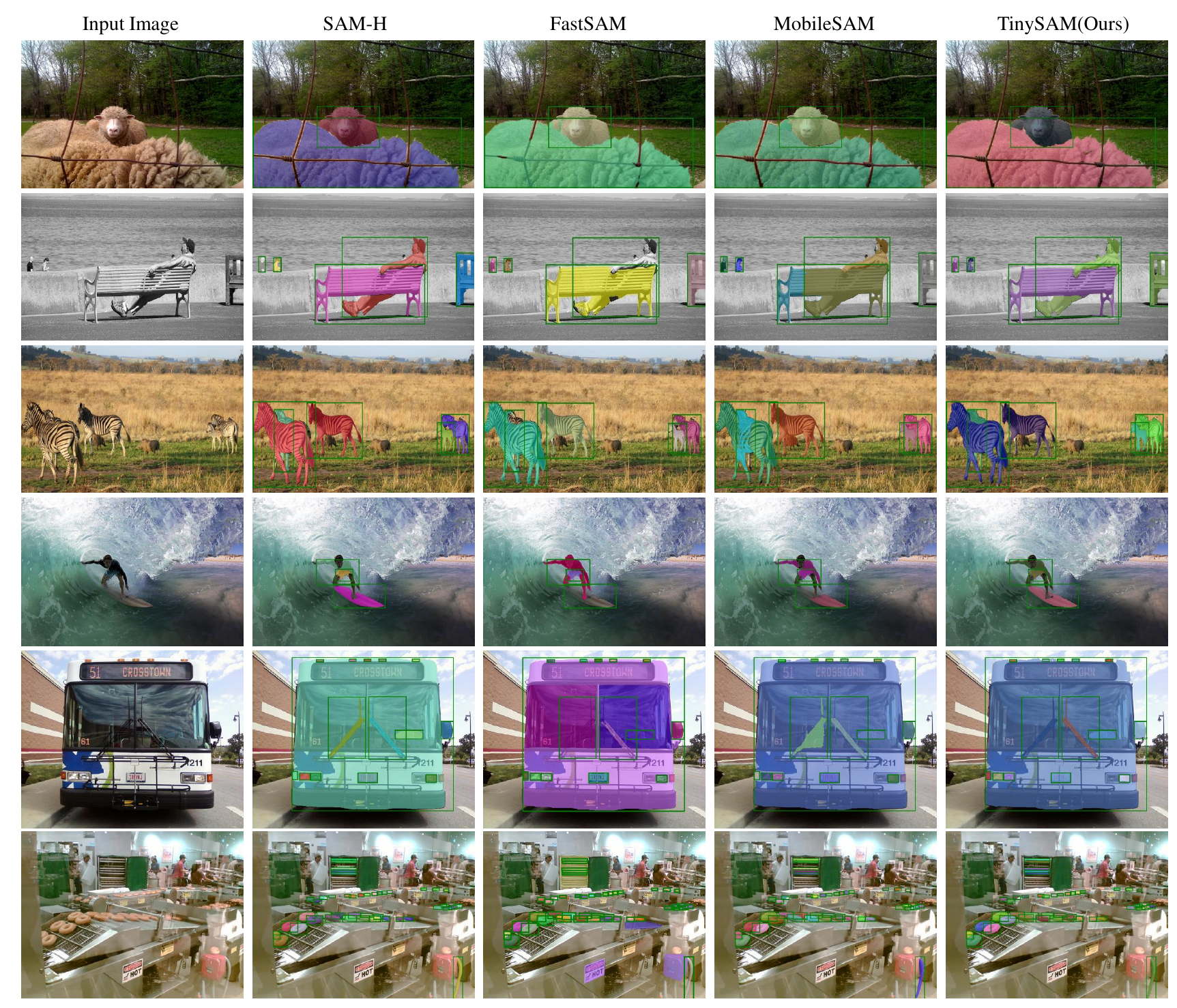}
		\vspace{-1em}
		\captionof{figure}{
			Visualization results of COCO validation dataset~(upper 3 rows) and LVIS v1 dataset~(lower 3 rows) for zero-shot instance segmentation. The green box marks the box prompt from the ViTDet-H detector. TinySAM captures more clear and smooth boundaries especially for hard targets of small size or similar texture feature.
		}
		\label{fig:box}
		\vspace{3em} 
	\end{center}%

}]

\section{Appendix}
We provide more visualization results for the appendix. Figure~\ref{fig:box} shows zero-shot instance segmentation on COCO dataset~\cite{lin2014microsoft} and LVIS v1~\cite{gupta2019lvis} dataset, respectively. For clear presentation, only detected boxes by ViTDet-H~\cite{vitdet} with higher confidence scores  than $0.8$ are prompted into models and visualized on the figure. LVIS v1 dataset has more fine-grained mask labels than COCO dataset~\cite{lin2014microsoft}, on which the proposed TinySAM demonstrates greater advantage in terms of both accuracy and efficiency.

\begin{figure*}[htp]
	\centering
	\vspace{0em}
	\includegraphics[width=1.0\linewidth]{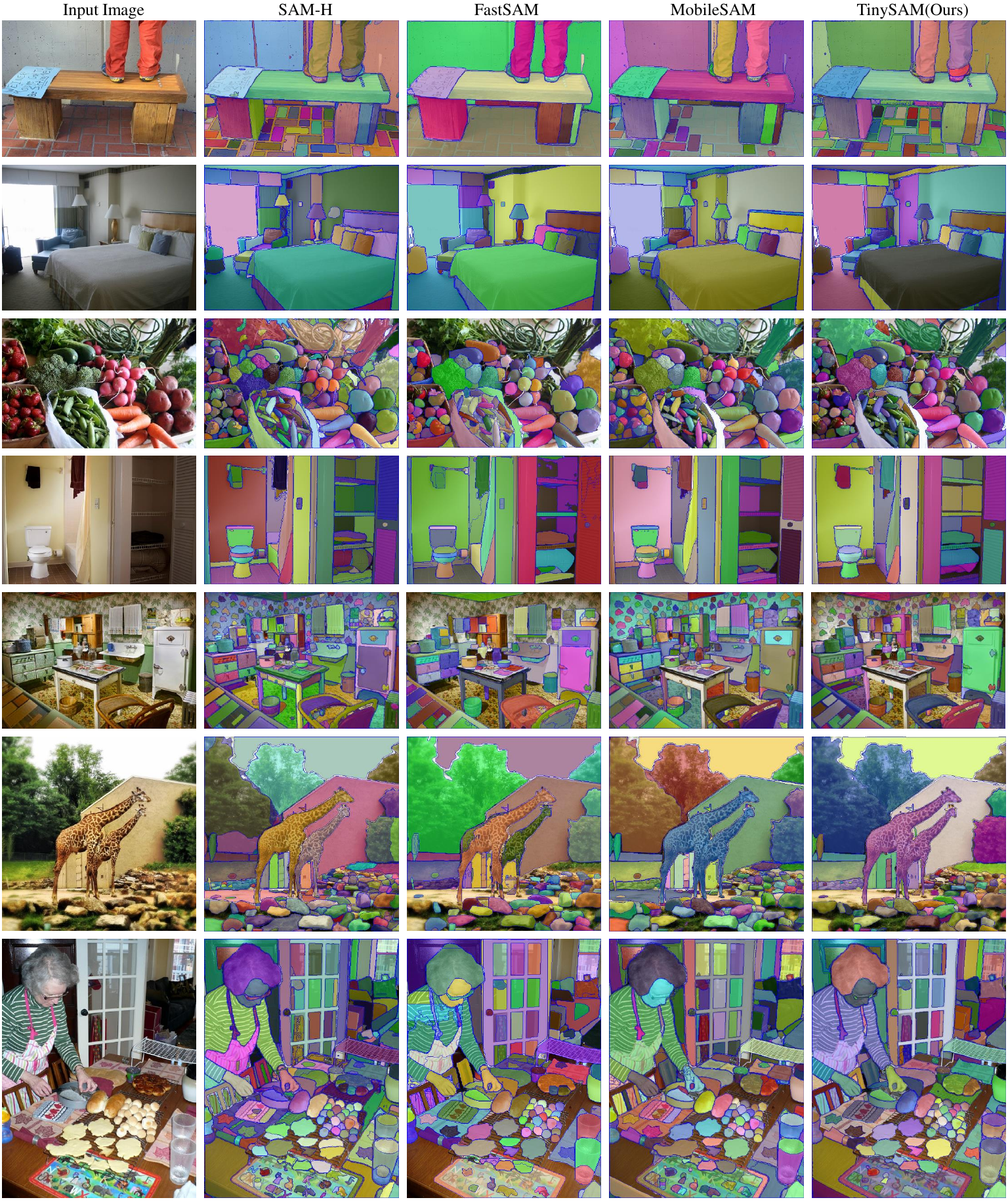}
	\vspace{0em}
	\caption{Visualization results of TinySAM model with hierarchical everything inference and its counterpart algorithms. Compared to FastSAM and MobileSAM, TinySAM captures fine-grained boundaries and masks, demonstrating similar performance with the computational expensive SAM-H model.}
	\label{fig:everything}
	\vspace{0mm}
\end{figure*}

Figure \ref{fig:everything} shows the everything inference results by the proposed TinySAM model with hierarchical everything inference and its counterpart algorithms. TinySAM captures clear boundaries and produces more fine-grained masks, whereas MobileSAM~\cite{mobilesam} and FastSAM~\cite{fastsam} sometimes generate fabricated boundaries and masks. TinySAM shows more close performance to the original SAM~\cite{SAM}, while consuming significantly less computation cost. %

\bibliography{aaai25}

\begin{thebibliography}{50}
\providecommand{\natexlab}[1]{#1}

\bibitem[{Bolya et~al.(2019)Bolya, Zhou, Xiao, and Lee}]{bolya2019yolact}
Bolya, D.; Zhou, C.; Xiao, F.; and Lee, Y.~J. 2019.
\newblock Yolact: Real-time instance segmentation.
\newblock In \emph{Proceedings of the IEEE/CVF international conference on
  computer vision}, 9157--9166.

\bibitem[{Caicedo et~al.(2019)Caicedo, Goodman, Karhohs, Cimini, Ackerman,
  Haghighi, Heng, Becker, Doan, McQuin et~al.}]{bbbc038v1}
Caicedo, J.~C.; Goodman, A.; Karhohs, K.~W.; Cimini, B.~A.; Ackerman, J.;
  Haghighi, M.; Heng, C.; Becker, T.; Doan, M.; McQuin, C.; et~al. 2019.
\newblock Nucleus segmentation across imaging experiments: the 2018 Data
  Science Bowl.
\newblock \emph{Nature methods}, 16(12): 1247--1253.

\bibitem[{Cen et~al.(2023)Cen, Zhou, Fang, Shen, Xie, Zhang, and
  Tian}]{cen2023segment3D}
Cen, J.; Zhou, Z.; Fang, J.; Shen, W.; Xie, L.; Zhang, X.; and Tian, Q. 2023.
\newblock Segment anything in 3d with nerfs.
\newblock \emph{arXiv preprint arXiv:2304.12308}.

\bibitem[{Chen et~al.(2017)Chen, Choi, Yu, Han, and Chandraker}]{chen2017kd}
Chen, G.; Choi, W.; Yu, X.; Han, T.; and Chandraker, M. 2017.
\newblock Learning efficient object detection models with knowledge
  distillation.
\newblock \emph{Advances in neural information processing systems}, 30.

\bibitem[{Chen et~al.(2020)Chen, Zhang, Wang, Shu, Xu, and
  Xu}]{chen2020optical}
Chen, X.; Zhang, Y.; Wang, Y.; Shu, H.; Xu, C.; and Xu, C. 2020.
\newblock Optical flow distillation: Towards efficient and stable video style
  transfer.
\newblock In \emph{Computer Vision--ECCV 2020: 16th European Conference,
  Glasgow, UK, August 23--28, 2020, Proceedings, Part VI 16}, 614--630.
  Springer.

\bibitem[{Chen et~al.(2024)Chen, Fang, Ma, and Wang}]{slimsam}
Chen, Z.; Fang, G.; Ma, X.; and Wang, X. 2024.
\newblock SlimSAM: 0.1\% Data Makes Segment Anything Slim.
\newblock In \emph{The Thirty-eighth Annual Conference on Neural Information
  Processing Systems}.

\bibitem[{Cheng et~al.(2022)Cheng, Misra, Schwing, Kirillov, and
  Girdhar}]{maskformer}
Cheng, B.; Misra, I.; Schwing, A.~G.; Kirillov, A.; and Girdhar, R. 2022.
\newblock Masked-attention mask transformer for universal image segmentation.
\newblock In \emph{Proceedings of the IEEE/CVF conference on computer vision
  and pattern recognition}, 1290--1299.

\bibitem[{Cheng et~al.(2023)Cheng, Li, Xu, Li, Yang, Wang, and
  Yang}]{cheng2023segment}
Cheng, Y.; Li, L.; Xu, Y.; Li, X.; Yang, Z.; Wang, W.; and Yang, Y. 2023.
\newblock Segment and track anything.
\newblock \emph{arXiv preprint arXiv:2305.06558}.

\bibitem[{Choi et~al.(2018)Choi, Wang, Venkataramani, Chuang, Srinivasan, and
  Gopalakrishnan}]{choi2018pact}
Choi, J.; Wang, Z.; Venkataramani, S.; Chuang, P. I.-J.; Srinivasan, V.; and
  Gopalakrishnan, K. 2018.
\newblock Pact: Parameterized clipping activation for quantized neural
  networks.
\newblock \emph{arXiv preprint arXiv:1805.06085}.

\bibitem[{Choukroun et~al.(2019)Choukroun, Kravchik, Yang, and
  Kisilev}]{choukroun2019low}
Choukroun, Y.; Kravchik, E.; Yang, F.; and Kisilev, P. 2019.
\newblock Low-bit quantization of neural networks for efficient inference.
\newblock In \emph{2019 IEEE/CVF International Conference on Computer Vision
  Workshop (ICCVW)}, 3009--3018. IEEE.

\bibitem[{Deng, Kong, and Murakami(2019)}]{Deng_2019_ICCV}
Deng, Z.; Kong, Q.; and Murakami, T. 2019.
\newblock Towards Efficient Instance Segmentation with Hierarchical
  Distillation.
\newblock In \emph{Proceedings of the IEEE/CVF International Conference on
  Computer Vision (ICCV) Workshops}.

\bibitem[{Dong et~al.(2023)Dong, Chen, Wang, and Xu}]{dong2023improving}
Dong, M.; Chen, X.; Wang, Y.; and Xu, C. 2023.
\newblock Improving Lightweight AdderNet via Distillation From $\ell_2$ to
  $\ell_1$-norm.
\newblock \emph{IEEE transactions on image processing: a publication of the
  IEEE Signal Processing Society}, 32: 5524--5536.

\bibitem[{Dosovitskiy et~al.(2020)Dosovitskiy, Beyer, Kolesnikov, Weissenborn,
  Zhai, Unterthiner, Dehghani, Minderer, Heigold, Gelly et~al.}]{vit}
Dosovitskiy, A.; Beyer, L.; Kolesnikov, A.; Weissenborn, D.; Zhai, X.;
  Unterthiner, T.; Dehghani, M.; Minderer, M.; Heigold, G.; Gelly, S.; et~al.
  2020.
\newblock An image is worth 16x16 words: Transformers for image recognition at
  scale.
\newblock \emph{arXiv preprint arXiv:2010.11929}.

\bibitem[{Esser et~al.(2019)Esser, McKinstry, Bablani, Appuswamy, and
  Modha}]{esser2019learned}
Esser, S.~K.; McKinstry, J.~L.; Bablani, D.; Appuswamy, R.; and Modha, D.~S.
  2019.
\newblock Learned step size quantization.
\newblock \emph{arXiv preprint arXiv:1902.08153}.

\bibitem[{Fortin et~al.(2022)Fortin, Gamache, Grondin, Pomerleau, and
  Gigu{\`e}re}]{timberseg}
Fortin, J.-M.; Gamache, O.; Grondin, V.; Pomerleau, F.; and Gigu{\`e}re, P.
  2022.
\newblock Instance segmentation for autonomous log grasping in forestry
  operations.
\newblock In \emph{2022 IEEE/RSJ International Conference on Intelligent Robots
  and Systems (IROS)}, 6064--6071. IEEE.

\bibitem[{Frantar et~al.(2022)Frantar, Ashkboos, Hoefler, and
  Alistarh}]{frantar2022gptq}
Frantar, E.; Ashkboos, S.; Hoefler, T.; and Alistarh, D. 2022.
\newblock Gptq: Accurate post-training quantization for generative pre-trained
  transformers.
\newblock \emph{arXiv preprint arXiv:2210.17323}.

\bibitem[{Guo et~al.(2021)Guo, Han, Wang, Wu, Chen, Xu, and
  Xu}]{guo2021distilling}
Guo, J.; Han, K.; Wang, Y.; Wu, H.; Chen, X.; Xu, C.; and Xu, C. 2021.
\newblock Distilling object detectors via decoupled features.
\newblock In \emph{Proceedings of the IEEE/CVF Conference on Computer Vision
  and Pattern Recognition}, 2154--2164.

\bibitem[{Gupta, Dollar, and Girshick(2019)}]{gupta2019lvis}
Gupta, A.; Dollar, P.; and Girshick, R. 2019.
\newblock Lvis: A dataset for large vocabulary instance segmentation.
\newblock In \emph{Proceedings of the IEEE/CVF conference on computer vision
  and pattern recognition}, 5356--5364.

\bibitem[{Hinton et~al.(2015)Hinton, Vinyals, Dean
  et~al.}]{hinton2015distilling}
Hinton, G.; Vinyals, O.; Dean, J.; et~al. 2015.
\newblock Distilling the knowledge in a neural network.
\newblock \emph{arXiv preprint arXiv:1503.02531}, 2(7).

\bibitem[{Jocher, Chaurasia, and Qiu(2023)}]{yolov8_ultralytics1}
Jocher, G.; Chaurasia, A.; and Qiu, J. 2023.
\newblock YOLO by Ultralytics.
\newblock \url{https://github.com/ultralytics/ultralytics}.

\bibitem[{Kirillov et~al.(2019)Kirillov, He, Girshick, Rother, and
  Doll{\'a}r}]{kirillov2019panoptic}
Kirillov, A.; He, K.; Girshick, R.; Rother, C.; and Doll{\'a}r, P. 2019.
\newblock Panoptic segmentation.
\newblock In \emph{Proceedings of the IEEE/CVF Conference on Computer Vision
  and Pattern Recognition}, 9404--9413.

\bibitem[{Kirillov et~al.(2023)Kirillov, Mintun, Ravi, Mao, Rolland, Gustafson,
  Xiao, Whitehead, Berg, Lo et~al.}]{SAM}
Kirillov, A.; Mintun, E.; Ravi, N.; Mao, H.; Rolland, C.; Gustafson, L.; Xiao,
  T.; Whitehead, S.; Berg, A.~C.; Lo, W.-Y.; et~al. 2023.
\newblock Segment anything.
\newblock In \emph{Proceedings of the IEEE/CVF International Conference on
  Computer Vision}, 4015--4026.

\bibitem[{Li et~al.(2022{\natexlab{a}})Li, Yang, Ji, Wang, and
  Cheng}]{li2022exploring}
Li, J.; Yang, T.; Ji, W.; Wang, J.; and Cheng, L. 2022{\natexlab{a}}.
\newblock Exploring denoised cross-video contrast for weakly-supervised
  temporal action localization.
\newblock In \emph{CVPR}, 19914--19924.

\bibitem[{Li et~al.(2022{\natexlab{b}})Li, Chen, Dong, Tang, Wang, and
  Xu}]{li2022spatial}
Li, Y.; Chen, X.; Dong, M.; Tang, Y.; Wang, Y.; and Xu, C. 2022{\natexlab{b}}.
\newblock Spatial-channel token distillation for vision mlps.
\newblock In \emph{International Conference on Machine Learning}, 12685--12695.
  PMLR.

\bibitem[{Li et~al.(2022{\natexlab{c}})Li, Mao, Girshick, and He}]{vitdet}
Li, Y.; Mao, H.; Girshick, R.; and He, K. 2022{\natexlab{c}}.
\newblock Exploring Plain Vision Transformer Backbones for Object Detection.
\newblock arXiv:2203.16527.

\bibitem[{Li et~al.(2022{\natexlab{d}})Li, Xu, Zhang, Cao, Gao, and
  Guo}]{li2022q}
Li, Y.; Xu, S.; Zhang, B.; Cao, X.; Gao, P.; and Guo, G. 2022{\natexlab{d}}.
\newblock Q-vit: Accurate and fully quantized low-bit vision transformer.
\newblock \emph{Advances in Neural Information Processing Systems}, 35:
  34451--34463.

\bibitem[{Lin et~al.(2017)Lin, Goyal, Girshick, He, and Doll{\'a}r}]{focal}
Lin, T.-Y.; Goyal, P.; Girshick, R.; He, K.; and Doll{\'a}r, P. 2017.
\newblock Focal loss for dense object detection.
\newblock In \emph{Proceedings of the IEEE international conference on computer
  vision}, 2980--2988.

\bibitem[{Lin et~al.(2014)Lin, Maire, Belongie, Hays, Perona, Ramanan,
  Doll{\'a}r, and Zitnick}]{lin2014microsoft}
Lin, T.-Y.; Maire, M.; Belongie, S.; Hays, J.; Perona, P.; Ramanan, D.;
  Doll{\'a}r, P.; and Zitnick, C.~L. 2014.
\newblock Microsoft coco: Common objects in context.
\newblock In \emph{Computer Vision--ECCV 2014: 13th European Conference,
  Zurich, Switzerland, September 6-12, 2014, Proceedings, Part V 13}, 740--755.
  Springer.

\bibitem[{Liu et~al.(2023)Liu, Niu, Yuan, Yang, Wang, and Liu}]{liu2023pd}
Liu, J.; Niu, L.; Yuan, Z.; Yang, D.; Wang, X.; and Liu, W. 2023.
\newblock Pd-quant: Post-training quantization based on prediction difference
  metric.
\newblock In \emph{CVPR}, 24427--24437.

\bibitem[{Liu et~al.(2018)Liu, Qi, Qin, Shi, and Jia}]{liu2018path}
Liu, S.; Qi, L.; Qin, H.; Shi, J.; and Jia, J. 2018.
\newblock Path aggregation network for instance segmentation.
\newblock In \emph{Proceedings of the IEEE conference on computer vision and
  pattern recognition}, 8759--8768.

\bibitem[{Liu et~al.(2019)Liu, Chen, Liu, Qin, Luo, and
  Wang}]{liu2019structured}
Liu, Y.; Chen, K.; Liu, C.; Qin, Z.; Luo, Z.; and Wang, J. 2019.
\newblock Structured knowledge distillation for semantic segmentation.
\newblock In \emph{Proceedings of the IEEE/CVF conference on computer vision
  and pattern recognition}, 2604--2613.

\bibitem[{Liu et~al.(2021{\natexlab{a}})Liu, Lin, Cao, Hu, Wei, Zhang, Lin, and
  Guo}]{liu2021swin}
Liu, Z.; Lin, Y.; Cao, Y.; Hu, H.; Wei, Y.; Zhang, Z.; Lin, S.; and Guo, B.
  2021{\natexlab{a}}.
\newblock Swin transformer: Hierarchical vision transformer using shifted
  windows.
\newblock In \emph{ICCV}, 10012--10022.

\bibitem[{Liu et~al.(2021{\natexlab{b}})Liu, Wang, Han, Zhang, Ma, and
  Gao}]{liu2021post}
Liu, Z.; Wang, Y.; Han, K.; Zhang, W.; Ma, S.; and Gao, W. 2021{\natexlab{b}}.
\newblock Post-training quantization for vision transformer.
\newblock \emph{Advances in Neural Information Processing Systems}, 34:
  28092--28103.

\bibitem[{Long, Shelhamer, and Darrell(2015)}]{fcnseg}
Long, J.; Shelhamer, E.; and Darrell, T. 2015.
\newblock Fully Convolutional Networks for Semantic Segmentation.
\newblock In \emph{Proceedings of the IEEE Conference on Computer Vision and
  Pattern Recognition (CVPR)}.

\bibitem[{Ma and Wang(2023)}]{ma2023segment}
Ma, J.; and Wang, B. 2023.
\newblock Segment anything in medical images.
\newblock \emph{arXiv preprint arXiv:2304.12306}.

\bibitem[{Milletari, Navab, and Ahmadi(2016)}]{dice}
Milletari, F.; Navab, N.; and Ahmadi, S.-A. 2016.
\newblock V-net: Fully convolutional neural networks for volumetric medical
  image segmentation.
\newblock In \emph{2016 fourth international conference on 3D vision (3DV)},
  565--571. IEEE.

\bibitem[{Nagel et~al.(2020)Nagel, Amjad, Van~Baalen, Louizos, and
  Blankevoort}]{nagel2020up}
Nagel, M.; Amjad, R.~A.; Van~Baalen, M.; Louizos, C.; and Blankevoort, T. 2020.
\newblock Up or down? adaptive rounding for post-training quantization.
\newblock In \emph{International Conference on Machine Learning}, 7197--7206.
  PMLR.

\bibitem[{Park et~al.(2019)Park, Kim, Lu, and Cho}]{Park_2019_CVPR}
Park, W.; Kim, D.; Lu, Y.; and Cho, M. 2019.
\newblock Relational Knowledge Distillation.
\newblock In \emph{Proceedings of the IEEE/CVF Conference on Computer Vision
  and Pattern Recognition (CVPR)}.

\bibitem[{Peng et~al.(2019)Peng, Jin, Liu, Li, Wu, Liu, Zhou, and
  Zhang}]{peng2019correlation}
Peng, B.; Jin, X.; Liu, J.; Li, D.; Wu, Y.; Liu, Y.; Zhou, S.; and Zhang, Z.
  2019.
\newblock Correlation congruence for knowledge distillation.
\newblock In \emph{Proceedings of the IEEE/CVF International Conference on
  Computer Vision}, 5007--5016.

\bibitem[{Pugliatti and Topputo(2022)}]{doors}
Pugliatti, M.; and Topputo, F. 2022.
\newblock DOORS: Dataset fOr bOuldeRs Segmentation.
\newblock \emph{Zenodo}, 9: 20.

\bibitem[{Romero et~al.(2014)Romero, Ballas, Kahou, Chassang, Gatta, and
  Bengio}]{romero2014fitnets}
Romero, A.; Ballas, N.; Kahou, S.~E.; Chassang, A.; Gatta, C.; and Bengio, Y.
  2014.
\newblock Fitnets: Hints for thin deep nets.
\newblock \emph{arXiv preprint arXiv:1412.6550}.

\bibitem[{Strudel et~al.(2021)Strudel, Garcia, Laptev, and Schmid}]{segmenter}
Strudel, R.; Garcia, R.; Laptev, I.; and Schmid, C. 2021.
\newblock Segmenter: Transformer for Semantic Segmentation.
\newblock In \emph{Proceedings of the IEEE/CVF International Conference on
  Computer Vision (ICCV)}, 7262--7272.

\bibitem[{Tai, Lin, and Wu(2023)}]{tai2023tsptq}
Tai, Y.-S.; Lin, M.-G.; and Wu, A.-Y.~A. 2023.
\newblock TSPTQ-ViT: Two-scaled post-training quantization for vision
  transformer.
\newblock In \emph{ICASSP 2023-2023 IEEE International Conference on Acoustics,
  Speech and Signal Processing (ICASSP)}, 1--5. IEEE.

\bibitem[{Wu et~al.(2020)Wu, Tang, Zhao, Zhang, Fu, and
  Zhang}]{wu2020easyquant}
Wu, D.; Tang, Q.; Zhao, Y.; Zhang, M.; Fu, Y.; and Zhang, D. 2020.
\newblock Easyquant: Post-training quantization via scale optimization.
\newblock \emph{arXiv preprint arXiv:2006.16669}.

\bibitem[{Wu et~al.(2022)Wu, Zhang, Peng, Liu, Xiao, Fu, and Yuan}]{tinyvit}
Wu, K.; Zhang, J.; Peng, H.; Liu, M.; Xiao, B.; Fu, J.; and Yuan, L. 2022.
\newblock Tinyvit: Fast pretraining distillation for small vision transformers.
\newblock In \emph{European Conference on Computer Vision}, 68--85. Springer.

\bibitem[{Xiong et~al.(2024)Xiong, Varadarajan, Wu, Xiang, Xiao, Zhu, Dai,
  Wang, Sun, Iandola et~al.}]{xiong2023efficientsam}
Xiong, Y.; Varadarajan, B.; Wu, L.; Xiang, X.; Xiao, F.; Zhu, C.; Dai, X.;
  Wang, D.; Sun, F.; Iandola, F.; et~al. 2024.
\newblock Efficientsam: Leveraged masked image pretraining for efficient
  segment anything.
\newblock In \emph{Proceedings of the IEEE/CVF Conference on Computer Vision
  and Pattern Recognition}, 16111--16121.

\bibitem[{Yu et~al.(2023)Yu, Feng, Feng, Liu, Jin, Zeng, and
  Chen}]{yu2023inpaint}
Yu, T.; Feng, R.; Feng, R.; Liu, J.; Jin, X.; Zeng, W.; and Chen, Z. 2023.
\newblock Inpaint anything: Segment anything meets image inpainting.
\newblock \emph{arXiv preprint arXiv:2304.06790}.

\bibitem[{Yuan et~al.(2022)Yuan, Xue, Chen, Wu, and Sun}]{yuan2022ptq4vit}
Yuan, Z.; Xue, C.; Chen, Y.; Wu, Q.; and Sun, G. 2022.
\newblock Ptq4vit: Post-training quantization for vision transformers with twin
  uniform quantization.
\newblock In \emph{ECCV}, 191--207. Springer.

\bibitem[{Zhang et~al.(2023)Zhang, Han, Qiao, Kim, Bae, Lee, and
  Hong}]{mobilesam}
Zhang, C.; Han, D.; Qiao, Y.; Kim, J.~U.; Bae, S.-H.; Lee, S.; and Hong, C.~S.
  2023.
\newblock Faster Segment Anything: Towards Lightweight SAM for Mobile
  Applications.
\newblock arXiv:2306.14289.

\bibitem[{Zhao et~al.(2023)Zhao, Ding, An, Du, Yu, Li, Tang, and
  Wang}]{fastsam}
Zhao, X.; Ding, W.; An, Y.; Du, Y.; Yu, T.; Li, M.; Tang, M.; and Wang, J.
  2023.
\newblock Fast Segment Anything.
\newblock arXiv:2306.12156.

\end{thebibliography}

\end{document}